\documentclass[conference]{IEEEtran}
\usepackage[utf8]{inputenc}
\usepackage{algorithm}
\usepackage{algpseudocode}
\usepackage{amsmath}
\usepackage{graphicx}
\usepackage{cite}
\usepackage{multicol}
\usepackage{hyperref}

\begin{document}

\title{Enhancing Sentiment Analysis in Bengali Texts: A Hybrid Approach Using Lexicon-Based Algorithm and Pretrained Language Model Bangla-BERT}
\author{
    \IEEEauthorblockN{\textbf{
        Hemal Mahmud\IEEEauthorrefmark{1},
        Hasan Mahmud\IEEEauthorrefmark{2},
        Mohammad Rifat Ahmmad Rashid\IEEEauthorrefmark{1}}
    }
    \vspace{0.2cm}
    \IEEEauthorblockA{
        \IEEEauthorrefmark{1}Department of Computer Science and Engineering, East West University, Dhaka, Bangladesh\\
        \IEEEauthorrefmark{2}Department of Computer Science and Engineering, Islamic University of Technology, Dhaka, Bangladesh}
}

\maketitle

\begin{abstract}
Sentiment analysis (SA) is a process of identifying the emotional tone or polarity within a given text and aims to uncover the user's complex emotions and inner feelings. While sentiment analysis has been extensively studied for languages like English, research in Bengali, remains limited, particularly for fine-grained sentiment categorization. This work aims to connect this gap by developing a novel approach that integrates rule-based algorithms with pre-trained language models. We developed a dataset from scratch, comprising over 15,000 manually labeled reviews. Next, we constructed a Lexicon Data Dictionary, assigning polarity scores to the reviews. We developed a novel rule based algorithm Bangla Sentiment Polarity Score (BSPS), an approach capable of generating sentiment scores and classifying reviews into nine distinct sentiment categories. To assess the performance of this method, we evaluated the classified sentiments using BanglaBERT, a pre-trained transformer-based language model. We also performed sentiment classification directly with BanglaBERT on the original data and evaluated this model's results. Our analysis revealed that the BSPS + BanglaBERT hybrid approach outperformed the standalone BanglaBERT model, achieving higher accuracy, precision, and nuanced classification across the nine sentiment categories. The results of our study emphasize the value and effectiveness of combining rule-based and pre-trained language model approaches for enhanced sentiment analysis in Bengali and suggest pathways for future research and application in languages with similar linguistic complexities.
\end{abstract}

\section{Introduction}
Sentiment analysis (SA) is a process of identifying the emotional tone or polarity within a given text and aimed to carefully untangle and uncover the user's complex emotions and inner feelings [2], delving into their psychological state to better understand and release the emotional layers that may be entangled or suppressed. Sentiment analysis and opinion mining is a research domain dedicated to understanding and interpreting people's opinions, emotions, attitudes, evaluations, and sentiments conveyed through written text [1]. It stands as a key area of investigation in natural language processing and is also widely explored within the fields of data mining, web mining, and text mining. In the era of technological progress and machine learning, a crucial aspect of Artificial Intelligence, it is both necessary and in high demand to extract emotions and contextual meanings from various sources, including newspapers, blogs, social media platforms, forum discussions, and opinions on specific posts. By analyzing these insights, businesses and organizations can better understand and predict user or customer behavior and patterns, allowing them to tailor products, services, and marketing strategies, improve customer experiences, and enhance decision-making processes.

Sentiment analysis of texts in English and other languages has been extensively studied using a wide variety of approaches. However, for Bangla, an ancient Indo-European language spoken by over 250 million people [3], research in sentiment analysis has been limited. While several machine learning, deep learning, rule-based, and pretrained language models have been applied to analyze sentiment in Bengali texts, few have achieved the desired results. Sentiment classification is often restricted to binary or ternary categories, despite the potential to detect a broader range of sentiment classes. There has been limited exploration of combined or hybrid methods for sentiment analysis in Bengali text. The intricate structure and semantic nuances of the Bengali language make rule-based or lexicon-based approaches highly sought after. A hybrid approach that combines lexicon-based data dictionaries [4] approach with pre-trained language models [5] is particularly rare for sentiment analysis in Bengali text. Given the proven effectiveness of pretrained language models, integrating this with lexicon-based approaches tailored to specific language domains could yield dynamic and impressive results.

In this paper, we first introduced a novel rule-based algorithm called the Bangla Sentiment Polarity Score (BSPS), specifically designed to analyze Bengali text. The BSPS algorithm generates sentiment scores for Bengali reviews, going beyond the conventional binary or ternary sentiment classifications typically used in sentiment analysis. Instead, it categorizes the sentiment into nine distinct classes, allowing for a more detailed and nuanced understanding of user emotions and feelings. This extended classification system helps better articulate the complexities of sentiment in Bengali text, providing a richer analysis of emotional tones expressed by users. Following the sentiment classification process, we evaluated the generated sentiment categories using Bangla-Bert, a pretrained BERT model specifically fine-tuned for the Bengali language. This evaluation step helped validate the performance of our rule-based algorithm by comparing the results with those produced by an advanced deep learning model, known for its ability to capture contextual meaning in text. In the second phase, we employed Bangla-BERT, a pretrained BERT model fine-tuned for the Bengali language, for sentiment classification. Using Bangla-BERT, we classified the sentiment of the Bengali text into same nine categories based on the model's understanding of context and linguistic nuances. Following this, we conducted an evaluation of the sentiment categories using the same Bangla-BERT model. This evaluation process allowed us to assess how accurately the model identified sentiment and whether it aligned with the expectations set by the initial rule-based classification. By comparing the results from both phases, we were able to explore the potential advantages of combining lexicon-based approaches (such as BSPS) with pretrained models like Bangla-BERT. This comparison allowed us to assess Bangla-BERT's performance in sentiment analysis, particularly when integrated with a lexicon-based method. The evaluation performed by Bangla-BERT provided valuable insights into the effectiveness of this hybrid approach.

\section{Related Work}
Due to the growing demand in recent decades, sentiment analysis has emerged as a critical focus for researchers in the ongoing development of AI. Many scholars have worked diligently to harness advancements in this field, concentrating on sentiment analysis across various languages, including not only English but also Arabic, Chinese, Hindi, French, Korean, and others. However, research in Bengali sentiment analysis remains limited, particularly in more advanced forms, mainly due to technical challenges, empirical constraints [6], and a lack of sufficient resources.

This paper's findings [7] have had a significant impact on our work, as, to the best of our knowledge, sentiment analysis in Bengali using an extended dictionary has not been explored in many researches. The authors of this paper developed a rule-based algorithm called Bangla Text Sentiment Score (BTSC), which, in conjunction with a lexicon data dictionary (LDD), was used to extract sentiments by generating sentiment scores. Subsequently, they applied supervised machine learning classifiers such as SVM, logistic regression (LR), and K-nearest neighbors (KNN) to further analyze the sentiment. In this paper [8], The authors proposed a method for Chinese text sentiment analysis that integrates both sentiment words and field-specific polysemic sentiment words, utilizing an expanded sentiment dictionary. The use of a lexicon-based data dictionary for the Arabic language has shown improved results in this paper compared to previous efforts.

In Akter and Aziz [9], the authors outlined a lexicon-based dictionary framework that determines sentiment by scrutinizing the frequency of emotional trigger words in each sentence. In M. M. H. Manik et al. [10], the authors manually compiled a balanced dataset of reviews across multiple sentiment categories. Their sentiment analysis model achieved high accuracy. However, the small size of the dataset and the uniform distribution of review categories might have influenced the results, leading to the high accuracy. In Alshari et al. [11], the authors described SentiWordNet (SWN) as being affected by the "curse of dimensionality," a problem that arises from the difficulties of handling high-dimensional data. To overcome this, they used a sentiment lexicon based on word2vec, a popular word embedding model, to perform sentiment analysis (SA), with the goal of enhancing the accuracy and efficiency of the analysis. In Zhang et al. [12], authors constructed an extended sentiment dictionary and A rule-based classifier was used to determine the polarity of the text by assigning a score to each sentence.
This study [13] compares the performance of BERT with Bi-LSTM, LSTM, and GRU models for sentiment classification and aspect detection on Bengali text. BERT outperforms the traditional models, demonstrating superior accuracy and highlighting the effectiveness of advanced NLP techniques for Bengali text analysis. This study [14] focuses on sentiment analysis of Bangla book reviews, addressing the gap in research on consumer sentiment in the e-commerce sector. The authors of this paper meticulously evaluated various deep neural network and transformer models, finding that XLM-R outperforms the others in terms of performance and accuracy at it's best. In the study [15], they explored various hybrid feature extraction techniques for Bangla Sentiment Analysis using a large dataset from microblogging platforms. The proposed method, combining Bangla-BERT with Skipgram, outperforms other techniques across machine learning, ensemble learning, and deep learning approaches, achieving superior accuracy in all areas. In the paper [16], they used lexicon-based approach and compare directly with ML models such as Decision Tree (DT), Naive Bayes (NB) and Support Vector Machine (SVM) classifiers. No pretrained model was used. This study [17] explores various Bengali sentiment analysis models, comparing traditional BiLSTM and BERT-based models across multiple datasets. It highlights the influence of dataset features on model performance and provides insights into the strengths of different models. This paper [18] presents the Low-Resource Team's approach for Task 2 of BLP-2023, focusing on SA of public social media posts and comments. They utilized BanglaBert, fine-tuned with various strategies and external datasets, and combined the best model variations into an ensemble.

\begin{figure*}[htbp]
            \centering
            \includegraphics[width=0.9\linewidth]{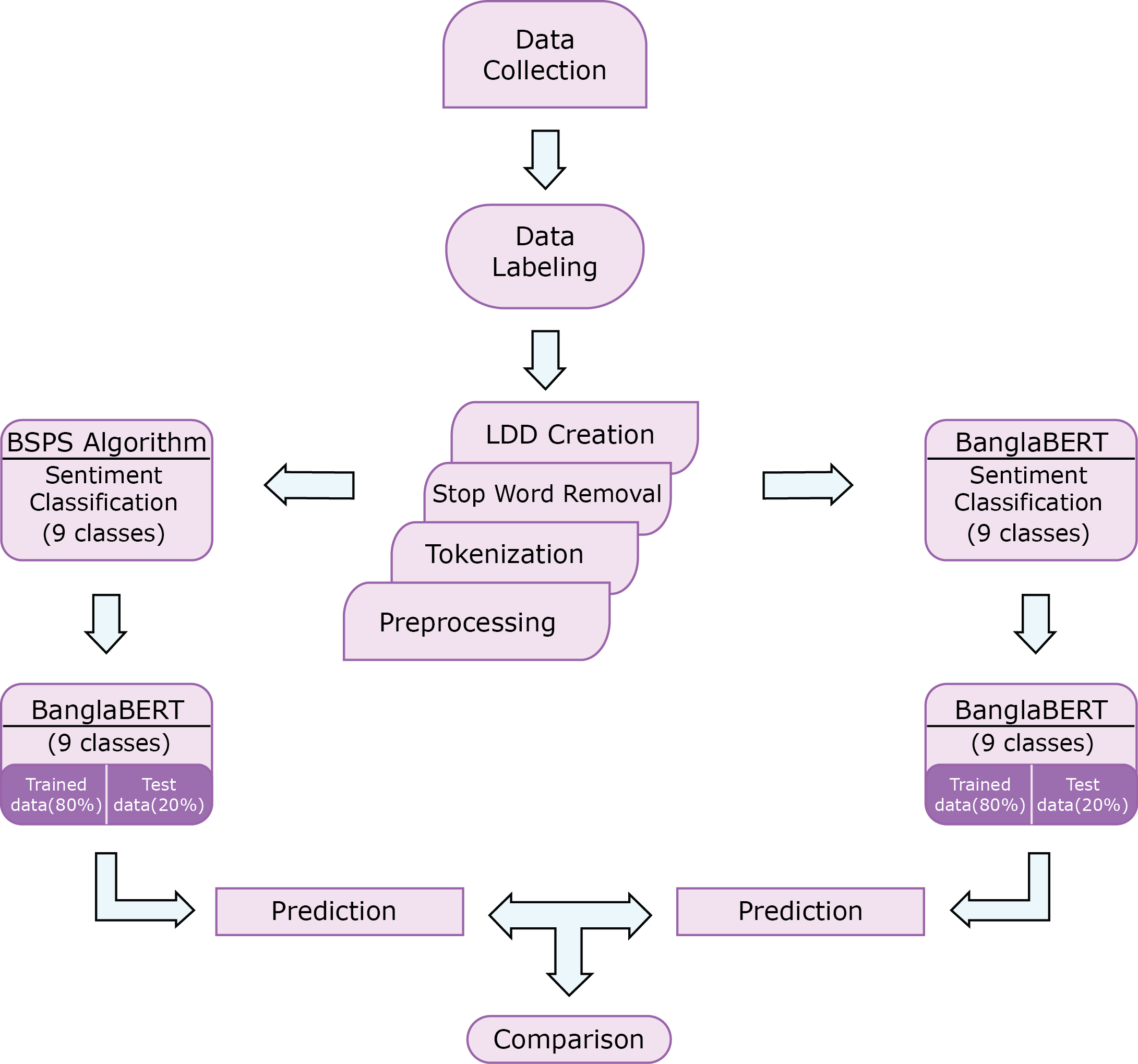}
            \caption{Visualization of methodology}
            \label{fig:performance}
\end{figure*}

\section{Methodology}
In this section, we provide a detailed description of the approach employed to carry out the study. As shown in \textbf{Fig.1}, our work is summarized at a glance. The methodology is structured in a series of key steps, which are outlined as follows:

\begin{enumerate}
    \vfill
    \item \textbf{Data Collection}:
    
    In the field of sentiment analysis, one of the major hurdles faced when working with the Bengali language is the scarcity of readily available and large-scale datasets for processing and analysis. Unlike English, where vast amounts of data are easily accessible for training models, Bengali suffers from a significant shortage of such datasets, particularly those with accurate sentiment labeling. While there are some small datasets available, they often lack the scale and quality needed for robust analysis. This limitation has made it exceedingly challenging to develop reliable sentiment analysis models for Bengali, especially those capable of handling diverse and complex linguistic features.

    Given the difficulty of acquiring sufficient data, we embarked on an extensive and resource-intensive process to manually collect and label a dataset specifically for this research. We painstakingly gathered a total of 15,194 instances of reviews from the Daraz Bangladesh website, a popular e-commerce platform. Out of these, 13,344 reviews were identified as positive, and 1,850 as negative. This manual data collection and labeling effort was crucial because it provided us with a high-quality dataset that was tailored to the requirements of Bengali sentiment analysis, compensating for the lack of existing, pre-labeled data. Through this laborious process, we were able to create a substantial dataset that could serve as a foundation for more accurate and comprehensive sentiment analysis models in the Bengali language.
    \vfill
    \item \textbf{LDD Creation}:
    
    The creation of the Lexicon Data Dictionary (LDD) was a key focus of our study, as it forms the core of our BSPS algorithm. The LDD was divided into two main categories: the positive lexicon and the negative lexicon. To construct these lexicons, we gathered a variety of positive and negative words with significant semantic meaning, which are crucial for sentiment analysis, from a widely used Kaggle resource [19].

    Once we assembled the initial positive and negative lexicons, we took an in-depth approach to expand them by adding various forms of each word, including different parts of speech and morphological variations that convey similar meanings. For instance, for the word 
    \raisebox{-0.09cm}{\includegraphics[width=0.06\textwidth]{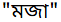}}
    [enjoyable] (noun), we conducted detailed research and included variations such as \raisebox{-0.09cm}{\includegraphics[width=0.07\textwidth]{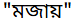}} [in a state of enjoyment] (noun), \raisebox{-0.09cm}{\includegraphics[width=0.07\textwidth]{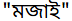}} [really enjoyable] (noun), \raisebox{-0.09cm}{\includegraphics[width=0.07\textwidth]{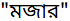}} [delicious] (adjective), and \raisebox{-0.09cm}{\includegraphics[width=0.075\textwidth]{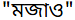}} [also tasty] (noun). This process was applied to all words of the dictionary in both lexicons to ensure comprehensive coverage of their meanings.

    After expanding the lexicons, we carefully analyzed each word and assigned a polarity score based on a structured and logical approach in accordance with the standards of the study [20]. We then normalized these scores, with a range of 0 to 1 for the positive lexicon [\textbf{TABLE 1}] and -1 to 0 for the negative lexicon [\textbf{TABLE 2}]. Following this, the LDD was ready for integration into the BSPS algorithm.

    \begin{table}[htbp]
    \vfill
    \caption{Positive Lexicon}
    \centering
    \renewcommand{\arraystretch}{2}
    \hspace{0.5 cm}
    \resizebox{0.4\textwidth}{!}{
    \begin{tabular}{|c|c|c|c|}
    \hline
    \textbf{Positive words} & \textbf{Polarity Score} \\
    \hline
    \raisebox{-0.05cm}{\includegraphics[width=0.045\textwidth]{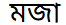}} & 0.7 \\
    \hline
    \raisebox{-0.09cm}{\includegraphics[width=0.08\textwidth]{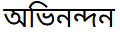}} & 0.6 \\
    \hline
    \raisebox{-0.05cm}{\includegraphics[width=0.05\textwidth]{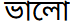}} & 0.9 \\
    \hline
    \raisebox{-0.05cm}{\includegraphics[width=0.065\textwidth]{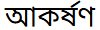}} & 0.7 \\
    \hline
    \raisebox{-0.05cm}{\includegraphics[width=0.075\textwidth]{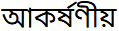}} & 0.7 \\
    \hline
    \raisebox{-0.05cm}{\includegraphics[width=0.085\textwidth]{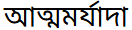}} & 0.4 \\
    \hline
    \end{tabular}
    }
    \end{table}

    \begin{table}[htbp]
    \vfill
    \caption{Negative Lexicon}
    \centering
    \renewcommand{\arraystretch}{2}
    \hspace{0.5 cm}
    \resizebox{0.4\textwidth}{!}{
    \begin{tabular}{|c|c|c|c|}
    \hline
    \textbf{Positive words} & \textbf{Polarity Score} \\
    \hline
    \raisebox{-0.05cm}{\includegraphics[width=0.065\textwidth]{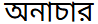}} & -0.6 \\
    \hline
    \raisebox{-0.05cm}{\includegraphics[width=0.033\textwidth]{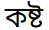}} & -0.8 \\
    \hline
    \raisebox{-0.05cm}{\includegraphics[width=0.055\textwidth]{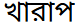}} & -0.9 \\
    \hline
    \raisebox{-0.05cm}{\includegraphics[width=0.04\textwidth]{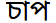}} & -0.5 \\
    \hline
    \raisebox{-0.05cm}{\includegraphics[width=0.047\textwidth]{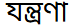}} & -0.8 \\
    \hline
    \raisebox{-0.07cm}{\includegraphics[width=0.04\textwidth]{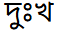}} & -0.85 \\
    \hline
    \end{tabular}
    }
    \end{table}
    \vfill
    \item \textbf{Data Preprocessing}:
    
    We gathered data from various sources including academic papers, surveys, and experiments.
        \begin{enumerate}
            \item \textbf{Null and Duplicate data handling}:
            
            We conducted a thorough inspection of the dataset to identify and address any potential issues such as missing (null) values or duplicate entries. During this process, we discovered several instances of duplicate data, which could have compromised the integrity and accuracy of our analysis. To ensure that the dataset was of high quality and contained only unique, valid entries, we carefully removed the duplicates. This step was crucial in refining the dataset, ensuring that it was clean, consistent, and well-prepared for subsequent analysis.
            
            \item \textbf{Tokenization and Normalization}:
            
                While processing the reviews, we employed regular expressions (regexp) to split the sentences into individual words, a process known as tokenization. This method allowed us to efficiently break down the text into smaller units, or tokens, which are essential for analyzing and understanding the sentiment expressed in each review. By utilizing regular expressions (regexp), we ensured a precise and effective segmentation of the text. For example: \raisebox{-0.05cm}{\includegraphics[width=0.18\textwidth]{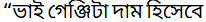}}\raisebox{-0.05cm}{\includegraphics[width=0.2\textwidth]{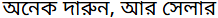}}, \raisebox{-0.05cm}{\includegraphics[width=0.195\textwidth]{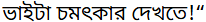}} [Brother, the shirt was very nice considering the price and the seller was good looking], when we tokenized this sentence, we got a list, as like [\raisebox{-0.05cm}{\includegraphics[width=0.05\textwidth]{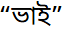}} [brother], \raisebox{-0.05cm}{\includegraphics[width=0.08\textwidth]{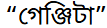}} [shirt], \raisebox{-0.05cm}{\includegraphics[width=0.052\textwidth]{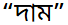}} [price], \raisebox{-0.05cm}{\includegraphics[width=0.07\textwidth]{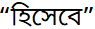}} [considering], \raisebox{-0.05cm}{\includegraphics[width=0.073\textwidth]{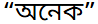}} [very], \raisebox{-0.05cm}{\includegraphics[width=0.066\textwidth]{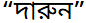}} [nice], \raisebox{-0.05cm}{\includegraphics[width=0.06\textwidth]{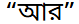}} [and], \raisebox{-0.05cm}{\includegraphics[width=0.07\textwidth]{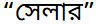}} [seller], \raisebox{-0.05cm}{\includegraphics[width=0.07\textwidth]{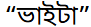}} [brother], \raisebox{-0.05cm}{\includegraphics[width=0.085\textwidth]{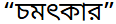}} [good], \raisebox{-0.05cm}{\includegraphics[width=0.075\textwidth]{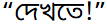}} [looking!]]. During the tokenization process, we also incorporated a normalization step. This step focused on cleaning the text by removing unnecessary characters, such as punctuation marks (e.g., ", ", ".", "!", "@", "\#", "\%"), which could potentially introduce noise into the analysis. In our example, [\raisebox{-0.05cm}{\includegraphics[width=0.075\textwidth]{w31.png}} [looking!]] changed to \raisebox{-0.05cm}{\includegraphics[width=0.075\textwidth]{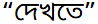}} [looking]]. By eliminating these extraneous symbols, we ensured that the text was in a more consistent and standardized form.
            
            \item \textbf{Stop Word Removal}:
            
            Many words, known as "stop words," play a role in sentence structure but do not contribute significant meaning, particularly in the context of sentiment analysis. We gathered a list of these "stop words" from [21] and incorporated them into our "Stop Word Removal" process to enhance the quality of our analysis. After this step, our above example look like this: [\raisebox{-0.05cm}{\includegraphics[width=0.05\textwidth]{w21.png}} [brother], \raisebox{-0.05cm}{\includegraphics[width=0.08\textwidth]{w22.png}} [shirt], \raisebox{-0.05cm}{\includegraphics[width=0.052\textwidth]{w23.png}} [price], \raisebox{-0.05cm}{\includegraphics[width=0.07\textwidth]{w24.png}} [considering], \raisebox{-0.05cm}{\includegraphics[width=0.073\textwidth]{w25.png}} [very], \raisebox{-0.05cm}{\includegraphics[width=0.066\textwidth]{w26.png}} [nice], 
            \raisebox{-0.05cm}{\includegraphics[width=0.07\textwidth]{w28.png}} [seller], \raisebox{-0.05cm}{\includegraphics[width=0.07\textwidth]{w29.png}} [brother], \raisebox{-0.05cm}{\includegraphics[width=0.085\textwidth]{w30.png}} [good], \raisebox{-0.05cm}{\includegraphics[width=0.075\textwidth]{w31.png}} [looking!]].. It can be seen that \raisebox{-0.05cm}{\includegraphics[width=0.06\textwidth]{w27.png}} [and] now doesn’t belong to tokens.
        \end{enumerate}

    \vfill
    \item \textbf{Explaination of BSPS Algorithm}:
    
    The Bangla Sentiment Polarity Score (BSPS) algorithm (\textbf{Algorithm 1}) is an advanced sentiment analysis algorithm is specifically designed for processing Bengali text which have semantic meaning and dialect complexities. Leveraging a combination of LDD and linguistic rules, \textbf{Algorithm 1} evaluates the sentiment of sentences analyzing key components such as positive and negative words, negation terms, extreme modifiers, phrase initiator etc. and by generating score for each review. The system is adept at handling the nuances of the Bengali language, including word order and the impact of conjunctions, phrases, and negations on sentiment. By utilizing a comprehensive lexicon, \textbf{Algorithm 1} attempts to provide an accurate sentiment score, helping to understand the underlying emotional tone of Bengali texts.
    \vfill
        \begin{enumerate}
            \item \textbf{Key Components}: 
                \begin{enumerate}
                    \item Positive Lexicons: Words that indicate positive sentiment (e.g., "\raisebox{-0.05cm}{\includegraphics[width=0.05\textwidth]{w8.png}}"  [good]). experiments.
                    \item Negative Lexicons: Words that indicate negative sentiment (e.g., "\raisebox{-0.05cm}{\includegraphics[width=0.05\textwidth]{w14.png}}" [bad]).
                    \item Direct Negation Words: Words that negate the sentiment (e.g., \raisebox{-0.05cm}{\includegraphics[width=0.042\textwidth]{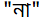}} [no]).
                    \item Extreme Words: Words that modify the sentiment intensity (e.g., \raisebox{-0.1cm}{\includegraphics[width=0.042\textwidth]{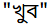}} [very], \raisebox{-0.07cm}{\includegraphics[width=0.07\textwidth]{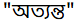}} [extremely]).
                    \item Phrases Starting Words: Certain words that is a starting of phrases (e.g., \raisebox{-0.05cm}{\includegraphics[width=0.12\textwidth]{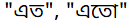}} [so or that much]) that might affect sentiment based on the context.

                    \item neg-flag: Indicates if a negation word is found, suggesting that the sentiment might be reversed.
                    \item pos-word-flag: Indicates if a positive word has been identified.
                    \item neg-word-flag: Indicates if a negative word has been identified.
                    \item extreme-flag: Indicates if an extreme word (like \raisebox{-0.064cm}{\includegraphics[width=0.042\textwidth]{w34.png}} [very]) has been identified, suggesting sentiment amplification.
                    \item phrase-flag: Tracks the presence of phrase-initial words that may modify the sentiment of the sentence.
                    \item pure-pos-flag / pure-neg-flag: Flags indicating whether the current sentiment is purely positive or negative, without modification by other factors.
                    \item and-flag: Used to track whether the word \raisebox{-0.05cm}{\includegraphics[width=0.15\textwidth]{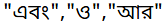}} [and or its equivalents] is present, which might join two positive or negative words together.
                    \item double-flag: Handles cases where two words are connected by word like \raisebox{-0.05cm}{\includegraphics[width=0.06\textwidth]{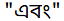}} [and or its equivalents].
                \end{enumerate}

            \item \textbf{Sentiment Processing Flow}:
            
                \begin{enumerate}
                    \item Tokenization and Filtering:
                    The input sentence is split into individual words (tokens) using a regular expression that captures word boundaries.
                    Stop words (common but unimportant words like \raisebox{-0.05cm}{\includegraphics[width=0.07\textwidth]{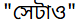}} [also that], \raisebox{-0.05cm}{\includegraphics[width=0.055\textwidth]{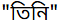}} [he]) are removed from the sentence.
                    \item Initialization:
                    \textbf{Algorithm 1} initializes a sentiment score (score = 0) and flags for tracking various conditions (e.g., neg-flag, pos-word-flag, etc.).
                    \item Processing Each Word:
                    \textbf{Algorithm 1} iterates through each word in the filtered sentence and processes it based on whether it's found in any of the predefined lexicons and lists (positive words, negative words, Direct Negation Words, Extreme Words etc.).
                    \item Handling "AND" Words:
                    If an "and-word" (like \raisebox{-0.05cm}{\includegraphics[width=0.062\textwidth]{w38.png}} [and]) is encountered, it triggers the and-flag. \textbf{Algorithm 1} looks at the previous and next words and applies a special rule for combining their sentiments (e.g., two positive words joined by "and" might amplify the sentiment).
                    \item Handling Phrases:
                    If a phrase-initial word (like \raisebox{-0.05cm}{\includegraphics[width=0.13\textwidth]{w36.png}} [that much]) is found, it triggers the phrase-flag. \textbf{Algorithm 1} then checks whether the word after the phrase initial is positive or negative and applies context-specific rules (e.g. \raisebox{-0.05cm}{\includegraphics[width=0.085\textwidth]{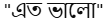}} [that much good] might indicate an intensified positive sentiment).
                    \item Handling Extreme Words:
                    If an extreme word is encountered, it triggers the extreme-flag. Extreme words amplify the sentiment score by a factor (e.g., \raisebox{-0.05cm}{\includegraphics[width=0.09\textwidth]{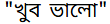}} [very good] becomes stronger than just "\raisebox{-0.05cm}{\includegraphics[width=0.045\textwidth]{w8.png}}" [good]). \textbf{Algorithm 1} checks if the extreme word is associated with a positive or negative word, applying the appropriate multiplier (e.g., \raisebox{-0.05cm}{\includegraphics[width=0.09\textwidth]{w42.png}} [very good] → score * 1.6).
                    \item Handling Negation:
                    When a negation word (like \raisebox{-0.05cm}{\includegraphics[width=0.09\textwidth]{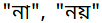}} [no or not]) is found, it triggers the neg-flag. \textbf{Algorithm 1} applies rules to reverse the sentiment of the associated word.
                    \item Double negation:
                    Double negation logic is also considered (e.g., \raisebox{-0.05cm}{\includegraphics[width=0.13\textwidth]{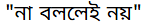}} [goes without saying] can negate the negation).
                    \item Final Calculation:
                    While processing all the words, \textbf{Algorithm 1} computes and updates the sentiment score based on each individual word it encounters, adjusted by rules for negation, conjunctions, and extremity.
                \end{enumerate}
            \vfill
            \item \textbf{Example Walk-through}:
            
                \begin{enumerate}
                    \item Example-01:
                    Given the sentence \raisebox{-0.05cm}{\includegraphics[width=0.22\textwidth]{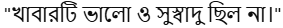}} [The food was not very good and delicious], here is how \textbf{Algorithm 1} processes it:

                    Tokenization:
                    
                    The sentence is split into tokens: \raisebox{-0.05cm}{\includegraphics[width=0.32\textwidth]{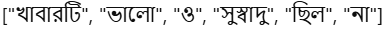}}.
                    Tokens after removing stop words: \raisebox{-0.05cm}{\includegraphics[width=0.28\textwidth]{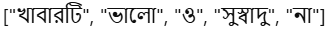}}.

                    Processing the tokens[\textbf{TABLE-3}]:

                    \begin{table}[htbp]
                    \vfill
                    \caption{Process of Example-01}
                    \centering
                    \renewcommand{\arraystretch}{2}
                    \hspace{0.6 cm}
                    \resizebox{0.45\textwidth}{!}{
                    \begin{tabular}{|c|c|c|c|}
                    \hline
                    \textbf{Token} & \textbf{Location} & \textbf{Score} & \textbf{Calculation} \\
                    \hline
                   \raisebox{-0.05cm}{\includegraphics[width=0.062\textwidth]{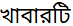}} & None & 0 & None \\
                    \hline
                    \raisebox{-0.05cm}{\includegraphics[width=0.053\textwidth]{w8.png}} & Positive-Lexicon & 0.9 & 0+0.9 \\
                    \hline
                    \raisebox{-0.05cm}{\includegraphics[width=0.03\textwidth]{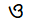}} & and-word & 0.9 & None \\
                    \hline
                    \raisebox{-0.05cm}{\includegraphics[width=0.052\textwidth]{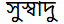}} & Positive-Lexicon & 1.6 & 0.9+0.7 \\
                    \hline
                    \raisebox{-0.05cm}{\includegraphics[width=0.035\textwidth]{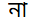}} & Direct Negation & -1.6 & 1.6*-1 \\
                    \hline
                    \end{tabular}
                    }
                    \end{table}
                    For the and-word "\raisebox{-0.05cm}{\includegraphics[width=0.03\textwidth]{w49.png}}" [and], we added the individual scores of "\raisebox{-0.05cm}{\includegraphics[width=0.05\textwidth]{w8.png}}" [good] and "\raisebox{-0.05cm}{\includegraphics[width=0.052\textwidth]{w50.png}}" [delicious] before applying negation technique for "\raisebox{-0.05cm}{\includegraphics[width=0.032\textwidth]{w51.png}}" [no]. Otherwise, the score would be like this: (+0.9) + (-0.7*-1) = 1.6, which is utterly contradictory and incorrect.
                    \vfill
                    \item Example-02: 
                    Given the sentence \raisebox{-0.05cm}{\includegraphics[width=0.22\textwidth]{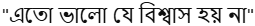}} [So good that it can't be believed], here’s how \textbf{Algorithm 1} processes it:

                    Tokenization:
                    The sentence is split into tokens: \raisebox{-0.05cm}{\includegraphics[width=0.31\textwidth]{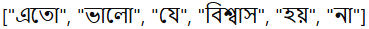}}.
                    
                    Tokens after removing stop words \raisebox{-0.05cm}{\includegraphics[width=0.075\textwidth]{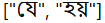}}: \raisebox{-0.05cm}{\includegraphics[width=0.24\textwidth]{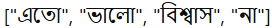}}.

                    Processing the tokens[\textbf{TABLE-4}]:
                    \begin{table}[htbp]
                    \caption{Process of Example-02}
                    \centering
                    \renewcommand{\arraystretch}{2}
                    \hspace{0.6 cm}
                    \resizebox{0.45\textwidth}{!}{
                    \begin{tabular}{|c|c|c|c|}
                        \hline
                        \textbf{Token} & \textbf{Location} & \textbf{Score} & \textbf{Calculation} \\
                        \hline
                        \raisebox{-0.05cm}{\includegraphics[width=0.05\textwidth]{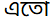}} & Phrase-Initial & 0 & None \\
                        \hline
                        \raisebox{-0.05cm}{\includegraphics[width=0.056\textwidth]{w8.png}} & Positive-Lexicon & 0.9 & 0 + 0.9 \\
                        \hline
                        \raisebox{-0.05cm}{\includegraphics[width=0.056\textwidth]{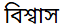}} & None & 0.9 & None \\ 
                        \hline
                        \raisebox{-0.05cm}{\includegraphics[width=0.034\textwidth]{w51.png}} & Direct Negation & 2.25 & 0.9 + 0.9 * 1.5 \\
                        \hline
                    \end{tabular} }
                    \end{table}
                    
                    When \textbf{Algorithm 1} encounters the word "\raisebox{-0.05cm}{\includegraphics[width=0.045\textwidth]{w56.png}}", it activates the phrase-flag. As a result, the direct negation word "\raisebox{-0.05cm}{\includegraphics[width=0.032\textwidth]{w51.png}}" does not reverse the sentiment but instead significantly amplifies the sentiment score. It makes sense as the sentence \raisebox{-0.05cm}{\includegraphics[width=0.22\textwidth]{w52.png}} [So good that it can't be believed] surely implies that the product is very very good.
                    \vfill
                    \item Example-03:
                    Given the sentence \raisebox{-0.05cm}{\includegraphics[width=0.21\textwidth]{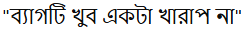}} [The bag is not very bad], here’s how the algorithm processes it:

                    Tokenization: The sentence is split into tokens: \raisebox{-0.05cm}{\includegraphics[width=0.28\textwidth]{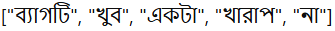}}.
                    
                    Tokens after removing stop words: \raisebox{-0.05cm}{\includegraphics[width=0.22\textwidth]{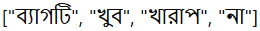}}.
                    
                    Processing the tokens[\textbf{TABLE-5}]:

                    \begin{table}[htbp] 
                    \vfill 
                    \caption{Process of Example-03} 
                    \centering 
                    \renewcommand{\arraystretch}{2} 
                    \hspace{0.6 cm} 
                    \resizebox{0.45\textwidth}{!}{ 
                        \begin{tabular}{|c|c|c|c|} 
                        \hline 
                        \textbf{Token} & \textbf{Location} & \textbf{Score} & \textbf{Calculation} \\ 
                        \hline 
                        \raisebox{-0.07cm}{\includegraphics[width=0.055\textwidth]{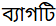}} & None & 0 & None \\ \hline 
                        \raisebox{-0.07cm}{\includegraphics[width=0.038\textwidth]{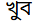}} & Extreme Word & 0 & None \\ 
                        \hline  
                        \raisebox{-0.07cm}{\includegraphics[width=0.061\textwidth]{w14.png}} & Negative Lexicon & -1.44 & -0.9 * 1.6 \\ 
                        \hline 
                        \raisebox{-0.07cm}{\includegraphics[width=0.038\textwidth]{w51.png}} & Direct Negation & 0.36 & -1.44 + (-0.9 * -2) \\ 
                        \hline 
                        \end{tabular} } 
                    \end{table}
                    \vspace{0.3cm}
                    In this case, we can see that, the sentence \raisebox{-0.07cm}{\includegraphics[width=0.21\textwidth]{w58.png}} [The bag is not very bad] essentially conveys the meaning of \raisebox{-0.07cm}{\includegraphics[width=0.17\textwidth]{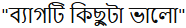}} [The bag is somewhat good]. Traditional rule-based algorithms typically did not account for this contextual interpretation. They would simply calculate \raisebox{-0.07cm}{\includegraphics[width=0.088\textwidth]{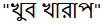}} [very bad], and then negate it due to the word "\raisebox{-0.07cm}{\includegraphics[width=0.034\textwidth]{w51.png}}" [not], resulting in \raisebox{-0.09cm}{\includegraphics[width=0.13\textwidth]{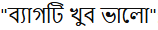}} [the bag is very good], which misrepresents the actual meaning of the sentence. However, our \textbf{Algorithm 1} carefully considers the subtle difference in meaning and incorporates this contextual nuance which is a coherent feature of Bengali language to produce the correct sentiment.
                    \vspace{0.2cm}
                \end{enumerate}
                
        \end{enumerate}
        \vspace{0cm}
        \begin{algorithm}
        \caption{BSPS ALGORITHM}
        \begin{algorithmic}[1]
        \vspace{0.3cm}
        \State \textbf{Input:} Sentence as a string
        \State \textbf{Output:} Sentiment score (float)
        
        \State Tokenize the sentence:
        \State Tokens $\gets \text{regexp\_tokenize}(sentence, \text{pattern} = "\backslash s|[\backslash.,!?;''']", \text{gaps=True})$
        \State Filtered\_tokens $\gets [word \text{ for word in } tokens \text{ if word} \notin \text{bengali\_stop\_words}]$
        
        \State Initialize: $score \gets 0$
        \State Initialize flags: All flags $\gets \text{False}$
        
        \For{each word in filtered\_tokens}
            \If{word in and\_word}
                \State $and\_flag \gets \text{True}$
            \ElsIf{word in phrase\_initial}
                \State $phrase\_flag \gets \text{True}$
            \ElsIf{word in extreme\_words}
                \State $extreme\_flag \gets \text{True}$
            \ElsIf{word in direct\_negation}
                \If{any flag is True}
                    \State Adjust score based on flags
                    \State Reset flags
                \Else
                    \State $neg\_flag \gets \text{True}$
                \EndIf
            \ElsIf{word in positive\_lexicon}
                \If{any flag is True}
                    \State Adjust score based on flags
                    \State Reset flags
                \Else
                    \State $pos\_word\_flag \gets \text{True}$
                    \State Update score
                \EndIf
            \ElsIf{word in negative\_lexicon}
                \If{any flag is True}
                    \State Adjust score based on flags
                    \State Reset flags
                \Else
                    \State $neg\_word\_flag \gets \text{True}$
                    \State Update score
                \EndIf
            \EndIf
        \EndFor
        
        \State \Return score
        \vfill
        \end{algorithmic}
        \end{algorithm}
        \vspace{1.5cm}

    \vspace{1.5cm}
    \vfill
    \item \textbf{Classification by BSPS}:
    
    Once we had generated sentiment scores for each review in our dataset using \textbf{Algorithm 1}, we undertook a thorough analysis of these scores. The raw sentiment scores were carefully evaluated to determine the emotional tone conveyed by each review. To ensure consistency and comparability across the reviews, we applied a normalization process to the scores.

    We first normalized the positive sentiment scores, scaling them to a range between 0 and 1, where 0 represents the most neutral sentiment, and 1 indicates the most positive sentiment. Similarly, the negative sentiment scores were normalized within the range of -1 to 0, where 0 indicates neutrality and -1 represents the most negative sentiment. This normalization process allowed for better clarity and consistency in interpreting the sentiment values, making them easier to compare and categorize.
    
    After normalization, we categorized the sentiment of each review into one of nine distinct classes. These categories are designed to represent varying degrees of positive, neutral, and negative sentiment. The categories are as follows:
        \begin{enumerate}
            \item Extremely Positive: Reviews with scores closest to 1, indicating a highly positive sentiment.
            \item Considerably Positive: Reviews with scores indicating strong positivity, but not as intense as "Extremely Positive."
            \item Positive: Reviews expressing a general positive sentiment.
            \item Slightly Positive: Reviews with a mild positive sentiment, showing a small preference for positive emotion.
            \item Neutral: Reviews that express no strong positive or negative emotions, often conveying a balanced sentiment.
            \item Slightly Negative: Reviews with a mild negative sentiment, indicating a small degree of dissatisfaction.
            \item Negative: Reviews with an overall negative sentiment.
            \item Considerably Negative: Reviews expressing strong dissatisfaction or negative emotion.
            \item Extremely Negative: Reviews with scores closest to -1, indicating highly negative sentiment.
        \end{enumerate}
        
        The purpose of this classification system was to provide a more detailed and nuanced understanding of user emotions. Instead of just classifying sentiment into broad categories like "positive" or "negative," this approach allowed us to better capture the range of emotions expressed in the reviews. By breaking down sentiment into finer categories, we can more accurately assess the intensity of user feelings, whether they are extremely positive, mildly negative, or somewhere in between. This categorization helps create a more granular sentiment analysis that can offer more precise insights into user opinions.
        
    \vspace{1cm}
    \item \textbf{BanglaBERT Hyperparameter Optimization on BSPS classification}:
    
    After creating the dataset with the nine sentiment categories, the next step was to fine-tune BanglaBERT, a transformer-based language model pre-trained on a large corpus of Bengali text. With the sentiment classes, referred to as "Advanced Sentiments," generated by our BSPS algorithm, we created a new dataset that included an additional "Advanced Sentiments" column. The steps that followed were as such:
        \begin{enumerate}
            \item \textbf{Training and Testing}: We divided the new dataset, which included the nine sentiment categories, into 80\% for training and 20\% for testing.
            \item \textbf{Preprocessing BanglaBERT}: Data preparation plays a crucial role in machine learning classification, as the performance of the model is largely determined by the quality of the input data [22]. For a transformer-based model like BanglaBERT, preprocessing ensures that the text is converted into numerical representations that the model can understand, maintaining the integrity and meaning of the original data. Each pre-processing task contributes to ensuring that the model can focus on learning meaningful patterns in the data, improving both its efficiency and accuracy during training and inference.

            Here are the main preprocessing tasks performed to prepare the Bengali text for input into BanglaBERT:
                \begin{enumerate}
                    \item Tokenization:
                    The text is divided into smaller, manageable units (tokens) using the BanglaBERT tokenizer to ensure the model can process the text.
                    \item Padding:
                    Sequences of different lengths are standardized by padding them to a fixed length, making sure all inputs have the same size.
                    \item Masking:
                    Padding tokens are masked so that the model ignores them during training, focusing only on the actual content of the text.
                    \item Label Encoding:
                    The sentiment labels are converted into integer values, making them compatible with the model’s output layer for classification tasks.
                \end{enumerate}
            \item \textbf{Fine-tuning BanglaBERT}: Fine-tuning a pre-trained model, such as BanglaBERT, is a challenging process that requires careful adjustment to ensure effective training.
                \begin{enumerate}
                    \item Hyper-parameters: During fine-tuning, we tested 10 different sets of learning rates (ranging from 1e-5 to 1e-3) and batch sizes (8, 16, and 32) to identify the optimal model performance.
                    \item Training Process: The model was fine-tuned over three epochs. We used the "Optuna" optimizer along with a dynamic learning rate scheduler and different batch sizes. The goal was for the model to map input reviews to one of the nine sentiment categories effectively.
                \end{enumerate}
                In the Fine-tuning BanglaBERT phase, we utilized 20\% of the dataset for each tuning iteration and collected predictions. Additionally, accuracy and other performance metrics were recorded at each step to evaluate and refine the model.
        \end{enumerate}
       
    \item \textbf{BanglaBERT language model for classification}:
    
    \begin{enumerate}
                    \item Model Selection: BanglaBERT was fine-tuned to classify reviews into positive and negative categories based on sentiment.
                    \item Fine-Tuning: The model was trained using 8 different sets of learning rate(ranging from 1e-5 to 1e-3) and batch size(8, 16, and 32) combinations to find the optimal parameters.
                    \item Probability Output: BanglaBERT generates a probability score that reflects the likelihood of a review being positive or negative.
                    \item Category Derivation: These scores were then sliced and mapped into one of the 9 sentiment categories.
                    \item Category Classification: Based on the probabilities, reviews were categorized into one of the 9 sentiment categories.
                \end{enumerate}

    \vspace{1cm}
    \item \textbf{BanglaBERT Hyperparameter Optimization on BanglaBERT classification}:
    
    After creating the dataset with the nine sentiment categories by fine tuning BanglaBERT, the next step was to evaluate by BanglaBERT. The steps that followed were as such:
        \begin{enumerate}
            \item \textbf{Training and Testing}: We again divided the new dataset, which included the nine sentiment categories, into 80\% for training and 20\% for testing.
            \item \textbf{Preprocessing BanglaBERT}:

            The pre-processing tasks performed to prepare the Bengali text for input into BanglaBERT:
                \begin{enumerate}
                    \item Tokenization:
                    The text is divided into smaller, manageable units (tokens) using the BanglaBERT tokenizer to ensure the model can process the text.
                    \item Padding:
                    Sequences of different lengths are standardized by padding them to a fixed length, making sure all inputs have the same size.
                    \item Masking:
                    Padding tokens are masked so that the model ignores them during training, focusing only on the actual content of the text.
                    \item Label Encoding:
                    The sentiment labels are converted into integer values, making them compatible with the model’s output layer for classification tasks.
                \end{enumerate}
            \item \textbf{Fine-tuning BanglaBERT}: 
            
                \begin{enumerate}
                    \item Hyperparameters: During fine-tuning, we tested 10 different sets of learning rates (ranging from 1e-5 to 1e-3) and batch sizes (8, 16, and 32) to identify the optimal model performance.
                    \item Training Process:
                    The model was fine-tuned over three epochs. We used the "Optuna" optimizer along with a dynamic learning rate scheduler and different batch sizes. The goal was for the model to effectively map input reviews to one of the nine sentiment categories.
                \end{enumerate}
                
                In the Fine-tuning BanglaBERT phase, we utilized 20\% of the dataset for each tuning iteration and collected predictions. Additionally, accuracy and other performance metrics were recorded at each step to evaluate and refine the model.
        \end{enumerate}
\end{enumerate}

The model was evaluated using cross-validation techniques to ensure robustness.

\section{Results}
To present the results of our work, we start by analyzing and evaluating the intermediate outcomes of the classification processes in both the BSPS and BanglaBERT models. In this stage, we predicted sentiments (limited to positive and negative as per the original dataset) for the entire dataset using: 1) the BSPS algorithm, which provides both sentiment labels and sentiment polarity scores, and 2) the pre-trained BanglaBERT model, which outputs probability values from which we inferred the sentiments for the entire dataset. We compare the performance of these two methods in predicting initial (positive and negative) sentiments. Following this, we examine the evaluation results and compare the two approaches used in our study: 1) sentiment classification with BSPS and evaluation with BanglaBERT, and 2) both sentiment classification and evaluation using BanglaBERT.

\begin{enumerate}
    \item \textbf{Evaluation metrics}:
        In our study, we utilized several evaluation metrics to assess the performance of our sentiment classification models, specifically focusing on precision, recall, and F1-score[\textbf{Fig.6}]. These metrics were calculated using the weighted average method, which takes into account the number of instances for each class. By applying the weighted average approach, we were able to ensure that the performance measures reflect the distribution of the classes in the dataset, giving more importance to classes with a higher number of instances.

        \begin{figure}[htbp]
            \centering
            \includegraphics[width=1.01\linewidth]{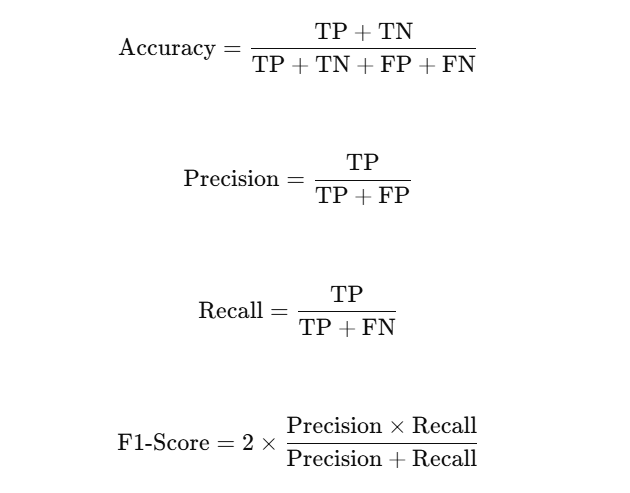}
            \caption{Evaluation metrics of the Model}
            \label{fig:performance}
        \end{figure}
        
    \vspace{0.2cm}    
    \item \textbf{Performance of the BSPS Algorithm}: 
        In our first approach of the BSPS algorithm for sentiment classification, the model achieved an impressive accuracy of 93\%, demonstrating its effectiveness in distinguishing between positive and negative sentiments. The algorithm produced both sentiment labels and polarity scores for the entire dataset, ensuring alignment with the original dataset's classification.  The BSPS algorithm performs well, showing that it’s both reliable and has the potential to be a valuable tool for analyzing sentiment in text.
        
        The following figures[\textbf{Fig.7, Fig.8}] illustrate the performance of the model:
        \begin{figure}[htbp]
            \centering
            \includegraphics[width=.99\linewidth]{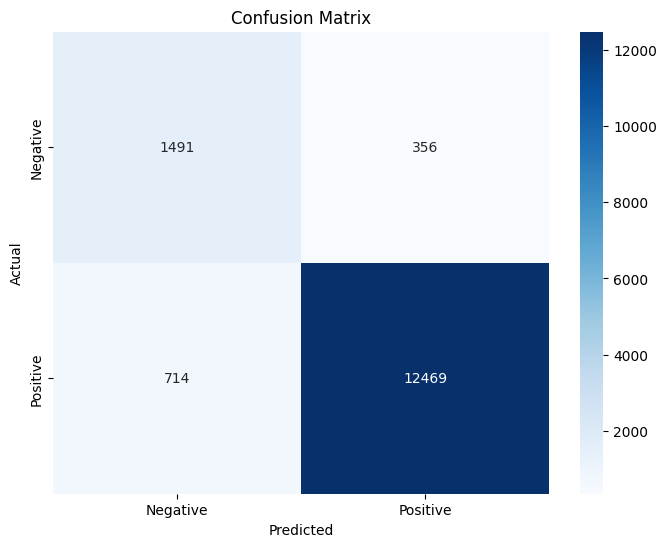}
            \caption{Confusion Matrix of Performance of the BSPS Algorithm}
            \label{fig:performance}
        \end{figure}

        \begin{figure}[htbp]
            \centering
            \includegraphics[width=.99\linewidth]{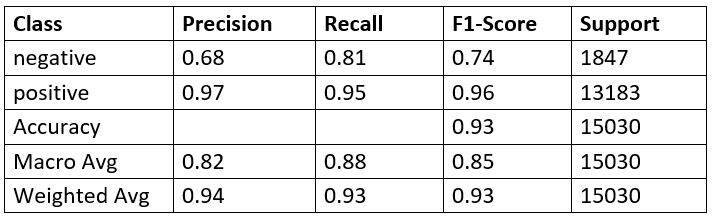}
            \caption{Performance Matrics of Performance of the BSPS Algorithm}
            \label{fig:performance}
        \end{figure}
        
    \vspace{0.2cm}    
    \item \textbf{Performance of the BanglaBERT (intermediate)}:
        In the intermediate phase of our study, we applied the BanglaBERT model for sentiment classification, focusing on positive and negative sentiment labels. The model achieved an accuracy of 88\%, showcasing its ability to correctly classify sentiment in a substantial portion of the dataset. Despite having some difficulties with unbalanced data, BanglaBERT produced precise sentiment labels and the corresponding polarity scores, which were analyzed to evaluate its effectiveness. Despite strong performance, the metrics indicate potential areas for improvement compared to the BSPS algorithm. Overall, BanglaBERT demonstrated solid performance, making it a valuable tool for sentiment analysis in the Bangla language.

        The following figures[\textbf{Fig.9, Fig.10}] illustrate the performance of the model:
        \begin{figure}[htbp]
            \centering
            \includegraphics[width=.99\linewidth]{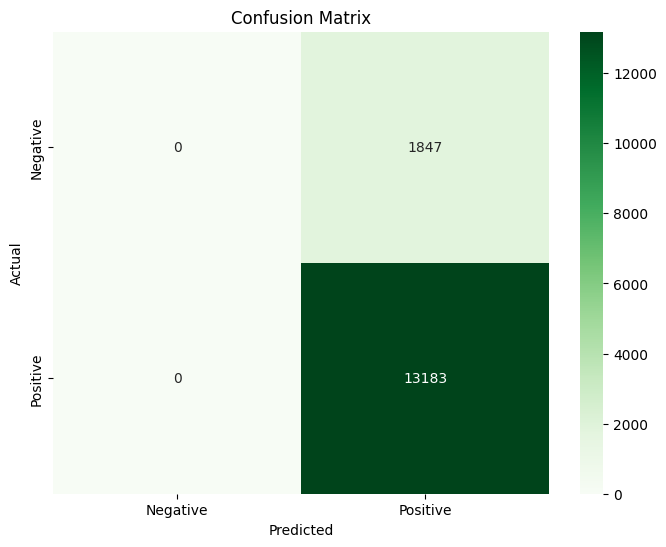}
            \caption{Confusion Matrix of Performance of BanglaBERT (intermediate)}
            \label{fig:performance}
        \end{figure}
        
        \begin{figure}[htbp]
            \centering
            \includegraphics[width=1.01\linewidth]{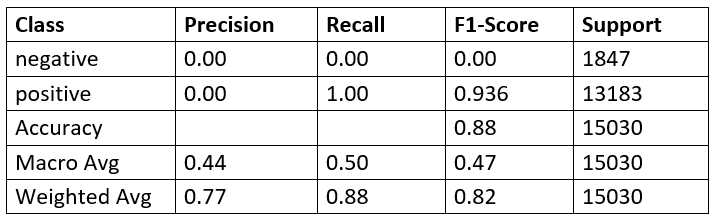}
            \caption{Performance Matrics of Performance of BanglaBERT (intermediate)}
            \label{fig:performance}
        \end{figure}
        \vspace{0.2cm}
        Although BanglaBERT is a powerful tool, it may not be as good at handling the subtle distinctions between closely related sentiment categories as BSPS. 
        The following figure[\textbf{Fig.11}] Illustrates the performance comparison between the two models at the intermediate level:
        \begin{figure}[htbp]
            \centering
            \includegraphics[width=1.01\linewidth]{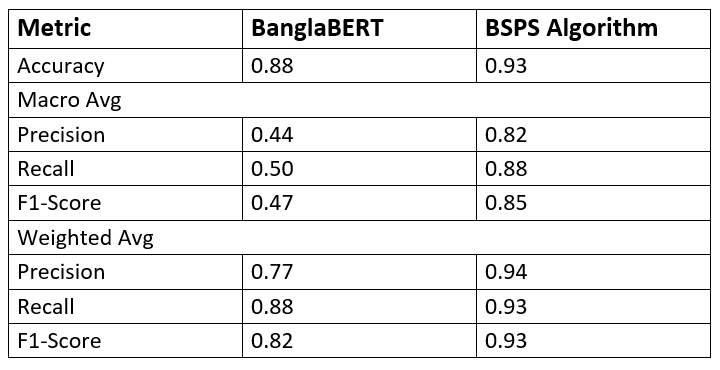}
            \caption{Comparison of Performance of BSPS and BanglaBERT (intermediate)}
            \label{fig:performance}
        \end{figure}
        
        \item \textbf{Performance Comparison: BSPS (classification) + BanglaBERT (evaluation) and BanglaBERT (classification) + BanglaBERT (evaluation)}:
        
        \vspace{0.2cm}
        The results indicate that BanglaBERT (classification) combined with BanglaBERT (evaluation) underperforms compared to BSPS (classification) + BanglaBERT (evaluation) across all evaluation metrics, including the accuracy (89\% vs. 79\%), weighted precision (0.89 vs. 0.69), weighted recall (0.89 vs. 0.79), and weighted F1-score (0.89 vs. 0.73). This suggests that the combination of using the BSPS algorithm for classification followed by BanglaBERT for evaluation yields better overall performance.
        For BSPS (classification) + BanglaBERT (evaluation): The best performance was achieved with the learning rate = 1.4e-5 and batch size = 32.
        For BanglaBERT (classification) + BanglaBERT (evaluation): The best performance was achieved with the learning rate = 1.8e-4 and batch size = 16. The lower learning rate (1.4e-5) of the BSPS algorithm and the larger batch size (32) likely contribute to more stable training and more accurate sentiment predictions. In contrast, the BanglaBERT-only pipeline, with a higher learning rate (1.8e-4) and smaller batch size (16), may have struggled with effective learning, which impacted its predictive performance. This comparison underscores the advantage of leveraging the BSPS algorithm for the initial classification step while using BanglaBERT for evaluation, although further optimization of BanglaBERT's hyperparameters could improve its results when used alone. 
        
        The \textbf{ Fig.12} illustrates the comparison of performances of the two models. 

        \vspace{0.2cm}
        \begin{figure}[htbp]
            \centering
            \includegraphics[width=0.99\linewidth]{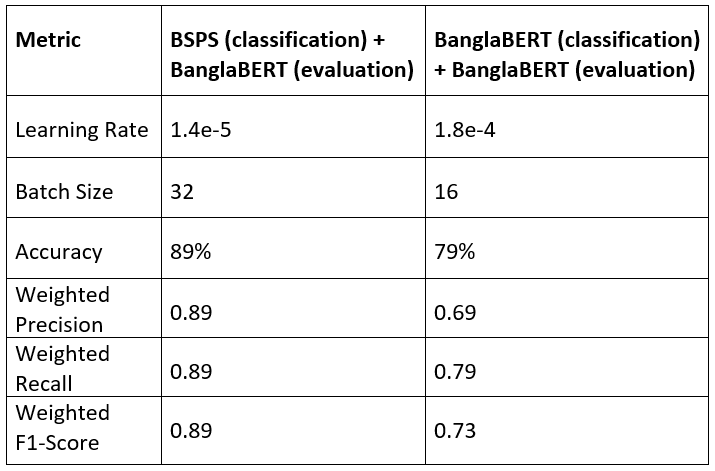}
            \caption{Final Performance comparison: BSPS+BanglaBERT vs BanglaBERT+BanglaBERT}
            \label{fig:performance}
        \end{figure}
\end{enumerate}
\vspace{1cm}
\section{Conclusion}
The results of this study reveal that our proposed model, which combines a rule-based algorithm with a pre-trained language model, offers significant advantages for sentiment analysis in Bengali. Specifically, Pipeline 1, which used the Bangla Sentiment Polarity Score (BSPS) algorithm followed by fine-tuning with BanglaBERT, achieved higher accuracy in classifying sentiment into 9 distinct categories compared to Pipeline 2, which used only the fine-tuned BanglaBERT model.

This highlights the effectiveness of the hybrid approach, where the BSPS algorithm efficiently preprocesses and categorizes reviews into broader sentiment categories, setting a strong foundation for the subsequent detailed analysis with BanglaBERT. Integration leads to better performance, faster learning, and more accurate results. Pipeline 2, while effective for binary classification tasks, showed limitations in handling complex multicategory sentiment analysis and may need further optimization for enhanced performance in nuanced sentiment classifications.

\vspace{1cm}
\section{Future Work}
Future improvements could involve experimenting with more sophisticated pre-trained models like mBERT or XLM-R, which might offer better performance by leveraging multilingual understanding and broader linguistic nuances. Furthermore, integrating the output of Pipeline 1 and Pipeline 2 through ensemble techniques could create a hybrid model that harnesses the strengths of both approaches, leading to more robust sentiment analysis across different types of data and scenarios.


\vspace{1cm}
\begin{thebibliography}{1}
\vspace{0.3cm}
\bibitem{Devika2016}
M.D. Devika, C. Sunitha, and A. Ganesh, \emph{Sentiment analysis: A comparative study on different approaches}, Procedia Computer Science, vol. 87, pp. 44-49, 2016.

\vspace{0.3cm}
\bibitem{Liu2012}
B. Liu, \emph{Sentiment Analysis and Opinion Mining}, Morgan \& Claypool Publishers LLC, 2012.

\vspace{0.3cm}
\bibitem{Devika2016b}
M.D. Devika, C. Sunitha, and A. Ganesh, \emph{Sentiment Analysis: A Comparative Study on Different Approaches}, Procedia Computer Science, vol. 87, pp. 44-49, 2016.

\vspace{0.3cm}
\bibitem{WikipediaBengali}
Wikipedia, \emph{Bengali language}, 

\url{https://en.wikipedia.org/wiki/Bengali_language}.

\vspace{0.3cm}
\bibitem{Taboada2011}
M. Taboada, J. Brooke, M. Tofiloski, K. Voll, and M. Stede, \emph{Lexicon-based methods for sentiment analysis}, Comput. Linguist., vol. 37, pp. 267-307, 2011.

\vspace{0.3cm}
\bibitem{Santhiya2024}
S. Santhiya, N. Abinaya, P. Jayadharshini, M. Dharshini, B. Kumar, and S. Kumar, \emph{A Comparative Analysis of Pretrained Models for Sentiment Analysis on Restaurant Customer Reviews (CAPM-SARCR)}, Communications in Computer and Information Science, 2024. DOI: \url{10.1007/978-3-031-58495-4_10}.

\vspace{0.3cm}
\bibitem{Karim2013}
M.A. Karim, \emph{Technical Challenges and Design Issues in Bangla Language Processing}, IGI Global, 2013.

\vspace{0.3cm}
\bibitem{Bhowmik2021}
N.R. Bhowmik, M. Arifuzzaman, M.R.H. Mondal, and M. Islam, \emph{Bangla text sentiment analysis using supervised machine learning with extended lexicon dictionary}, Nat. Lang. Process. Res., vol. 1, pp. 34-45, 2021.

\vspace{0.3cm}
\bibitem{Xu2019}
G. Xu, Z. Yu, H. Yao, F. Li, Y. Meng, and X. Wu, \emph{Chinese text sentiment analysis based on extended sentiment dictionary}, IEEE Access, vol. 7, pp. 43749-43762, 2019.

\vspace{0.3cm}
\bibitem{Akter2016}
S. Akter and M.T. Aziz, \emph{Sentiment analysis on Facebook group using lexicon based approach}, in \emph{2016 3rd International Conference on Electrical Engineering and Information Communication Technology (ICEEICT)}, IEEE, Dhaka, Bangladesh, 2016, pp. 1-4.

\vspace{0.3cm}
\bibitem{Manik2022}
M.M.H. Manik et al., \emph{A Hybrid Framework for Sentiment Analysis from Bangla Texts}, in \emph{2022 25th International Conference on Computer and Information Technology (ICCIT)}, IEEE, 2022, pp. 1-6. DOI: \url{10.1109/ICCIT57492.2022.10054952}.

\vspace{0.3cm}
\bibitem{Alshari2018}
E.M. Alshari, A. Azman, S. Doraisamy, N. Mustapha, and M. Alkeshr, \emph{Effective method for sentiment lexical dictionary enrichment based on word2vec for sentiment analysis}, in \emph{2018 Fourth International Conference on Information Retrieval and Knowledge Management (CAMP)}, IEEE, Kota Kinabalu, Malaysia, 2018, pp. 1-5.

\vspace{0.3cm}
\bibitem{Zhang2019}
Zhang et al., \emph{Chinese text sentiment analysis based on extended sentiment dictionary}, IEEE Access, vol. 7, pp. 43749-43762, 2019.

\vspace{0.3cm}
\bibitem{Junayed2023}
H. Junayed, M. Sheikh, M. Abdullah, B. Mohammad, F. Monir, \emph{Exploring the Efficacy of BERT in Bengali NLP: A Study on Sentiment Analysis and Aspect Detection}, pp. 54-59, 2023. DOI: \url{10.1109/eurocon56442.2023.10199048}.

\vspace{0.3cm}
\bibitem{Gobinda2022}
G.C. Sarker, K. Faysal, M.A. Sadat, A.R. Das, \emph{Book Review Sentiment Classification in Bangla using Deep Learning and Transformer Model}, IEEE, 2022. DOI: \url{10.1109/STI56238.2022.10103337}.

\vspace{0.3cm}
\bibitem{Shymon2023}
M.S. Islam, K.M. Alam, \emph{An Empiric Study on Bangla Sentiment Analysis Using Hybrid Feature Extraction Techniques}, IEEE, 2023. DOI: \url{10.1109/icccnt56998.2023.10308114}.

\vspace{0.3cm}
\bibitem{Dey2019}
R.C. Dey and O. Sarker, \emph{Sentiment analysis on bengali text using lexicon-based approach}, in \emph{2019 22nd International Conference on Computer and Information Technology (ICCIT)}, 2019. DOI: \url{10.1109/ICCIT48885.2019.9038250}.

\vspace{0.3cm}
\bibitem{Motahar2023}
M. Hossain, I. Ara, H. Akhi, R. Sama, T. Onni, A. Islam, \emph{Critical Analysis of BERT and LSTM Model for Bengali Sentiment Analysis Across Varied Datasets}, 2023. DOI: \url{10.1109/cicn59264.2023.10402353}.

\vspace{0.3cm}
\bibitem{Aunabil2023}
A. Chakma and M. Hasan, \emph{LowResource at BLP-2023 Task 2: Leveraging BanglaBert for Low Resource Sentiment Analysis of Bangla Language}, abs/2311.12735, 2023. DOI: \url{10.48550/arxiv.2311.12735}.

\vspace{0.3cm}
\bibitem{Nielsen2011}
Finn Årup Nielsen, "A new ANEW: evaluation of a word list for sentiment analysis in microblogs", Proceedings of the ESWC2011 Workshop on 'Making Sense of Microposts': Big things come in small packages. Volume 718 in CEUR Workshop Proceedings: 93-98. 2011 May. Matthew Rowe, Milan Stankovic, Aba-Sah Dadzie, Mariann Hardey (editors)

\vspace{0.3cm}
\bibitem{Akther2022}
Akther, A., Islam, M.S., Sultana, H., Rahman, A.R., Saha, S., Alam, K.M., Debnath, R.,2022. Compilation, analysis and application of a comprehensive bangla corpuskumono. IEEE Access 10, 79999–80014. http://dx.doi.org/10.1109/ACCESS.2022.3195236.

\vspace{0.3cm}
\bibitem{Hohman2020}
Hohman, F.; Wongsuphasawat, K.; Kery, M.B.; Patel, K. Understanding and visualizing data iteration in machine learning. In Proceedings of the 2020 CHI Conference on Human Factors in Computing Systems, Online, 25–30 April 2020; pp. 1–13

\end{thebibliography}
\end{document}